\documentclass[10pt,twocolumn,letterpaper]{article}

\usepackage[pagenumbers]{cvpr} 

\usepackage{graphicx}
\usepackage{amsmath}
\usepackage{amssymb}
\usepackage{booktabs}

\usepackage{nicefrac} 
\usepackage{xcolor}
\usepackage{graphicx}    
\usepackage{xspace}
\usepackage{tabularx}
\usepackage{multirow}
\usepackage{verbatim}
\usepackage{tabularx}
\usepackage{colortbl}
\usepackage{setspace}
\usepackage{array, makecell} %
\usepackage{xstring}

\usepackage{xspace}
\usepackage[export]{adjustbox}

\usepackage{bold-extra}

\makeatletter

\newcommand{\PAR}[1]{\vskip4pt \noindent {\bf #1~}}

\newcommand{\TODO}[1]{\textcolor{red}{#1}}

\definecolor{better_blue}{HTML}{4285f4}

\newtoggle{comments}
\toggletrue{comments}
\togglefalse{comments}

\iftoggle{comments}{%
\newcommand{\ali}[1]{\textcolor{blue}{\textbf{Ali: }{#1}}}
\newcommand{\alex}[1]{\textcolor[rgb]{0,0.5,0}{\textbf{Alex: }{#1}}}
\newcommand{\jono}[1]{\textcolor{brown}{\textbf{Jono: }{#1}}}
\newcommand{\deva}[1]{\textcolor[rgb]{0,0.5,0.5}{\textbf{Deva: }{#1}}}
\newcommand{\bastian}[1]{\textcolor[rgb]{0.5,0.5,0}{\textbf{Bastian: }{#1}}}
\newcommand{\ext}[1]{\textcolor{orange}{\textbf{Ext. Rev: }{#1}}}
\newcommand{\todo}[1]{\textcolor{red}{\small Todo:\,#1}\PackageWarning{TODO:}{#1!}}

}
{%
\newcommand{\ali}[1]{}
\newcommand{\alex}[1]{}
\newcommand{\jono}[1]{}
\newcommand{\deva}[1]{}
\newcommand{\bastian}[1]{}
\newcommand{\ext}[1]{}
\newcommand{\todo}[1]{}

 \setlength{\floatsep}{10pt plus2pt minus2pt}
 \setlength{\textfloatsep}{10pt plus2pt minus2pt}
 \setlength{\dblfloatsep}{10pt plus2pt minus2pt}
 \setlength{\dbltextfloatsep}{10pt plus2pt minus2pt}
 \addtolength{\abovecaptionskip}{-5pt}

}

\newcolumntype{P}[1]{>{\centering\arraybackslash}p{#1}}

\usepackage{pifont}

\newcommand{\J}{\mathcal{J}}
\newcommand{\F}{\mathcal{F}}
\newcommand{\JnF}{\mathcal{J}\&\mathcal{F}}

\newcommand{\real}[1]{\mathbb{R}^{#1}}
\newcommand{\embedding}[2]{\textbf{d}_{#1}^{#2}}

\newcommand{\smalltt}[1]{{\small\texttt{#1}}}

\makeatletter
\@namedef{ver@everyshi.sty}{}
\makeatother
\usepackage{csvsimple}
\usepackage{ifthen}

\newcommand{\cm}[1]{\ifthenelse{\equal{y}{#1}}{\checkmark}{\ifthenelse{\equal{n}{#1}}{}{???}}}
\newcommand{\hq}[2]{\IfSubStr{b}{#2}{\textbf{#1}}{#1}}

\newcommand{\tableline}[2]{
    \csvreader[
        column count=37,
        late after line=\\,
        filter=\equal{\key}{#1}
    ]{tables/data.csv}{
         1=\key,
         2=\fff,
         3=\ytvos,
         4=\framerate,
         5=\backbone,
         6=\GDavisVal, 7=\GDavisValH,
         8=\JDavisVal, 9=\JDavisValH,
        10=\FDavisVal, 11=\FDavisValH,
        12=\GDavisTest, 13=\GDavisTestH,
        14=\JDavisTest, 15=\JDavisTestH,
        16=\FDavisTest, 17=\FDavisTestH,
        18=\GYTVOS, 19=\GYTVOSH,
        20=\JYTseen, 21=\JYTseenH,
        22=\JYTunseen, 23=\JYTunseenH,
        24=\FYTseen, 25=\FYTseenH,
        26=\FYTunseen, 27=\FYTunseenH,
        28=\GYTVOSNew, 29=\GYTVOSNewH,
        30=\JYTNewseen, 31=\JYTNewseenH,
        32=\JYTNewunseen, 33=\JYTNewunseenH,
        34=\FYTNewseen, 35=\FYTNewseenH,
        36=\FYTNewunseen, 37=\FYTNewunseenH
    }{
        & #2 & \cm{\fff} & \hq{\GDavisVal}{\GDavisValH} & \hq{\JDavisVal}{\JDavisValH} & \hq{\FDavisVal}{\FDavisValH} && \hq{\GDavisTest}{\GDavisTestH} & \hq{\JDavisTest}{\JDavisTestH} & \hq{\FDavisTest}{\FDavisTestH} && \hq{\GYTVOSNew}{\GYTVOSNewH} & \hq{\JYTNewunseen}{\JYTNewunseenH} & \hq{\FYTNewunseen}{\FYTNewunseenH} & \hq{\JYTNewseen}{\JYTNewseenH} & \hq{\FYTNewseen}{\FYTNewseenH}
    }
}

\newcommand{\hqg}[2]{\textcolor{gray}{\IfSubStr{b}{#2}{\textbf{#1}}{#1}}}

\newcommand{\tablelineold}[2]{
    \csvreader[
        column count=37,
        late after line=\\,
        filter=\equal{\key}{#1}
    ]{tables/data.csv}{
         1=\key,
         2=\fff,
         3=\ytvos,
         4=\framerate,
         5=\backbone,
         6=\GDavisVal, 7=\GDavisValH,
         8=\JDavisVal, 9=\JDavisValH,
        10=\FDavisVal, 11=\FDavisValH,
        12=\GDavisTest, 13=\GDavisTestH,
        14=\JDavisTest, 15=\JDavisTestH,
        16=\FDavisTest, 17=\FDavisTestH,
        18=\GYTVOS, 19=\GYTVOSH,
        20=\JYTseen, 21=\JYTseenH,
        22=\JYTunseen, 23=\JYTunseenH,
        24=\FYTseen, 25=\FYTseenH,
        26=\FYTunseen, 27=\FYTunseenH,
        28=\GYTVOSNew, 29=\GYTVOSNewH,
        30=\JYTNewseen, 31=\JYTNewseenH,
        32=\JYTNewunseen, 33=\JYTNewunseenH,
        34=\FYTNewseen, 35=\FYTNewseenH,
        36=\FYTNewunseen, 37=\FYTNewunseenH
    }{
        & #2 & \cm{\fff} & \hq{\GDavisVal}{\GDavisValH} & \hq{\JDavisVal}{\JDavisValH} & \hq{\FDavisVal}{\FDavisValH} && \hq{\GDavisTest}{\GDavisTestH} & \hq{\JDavisTest}{\JDavisTestH} & \hq{\FDavisTest}{\FDavisTestH} && \hqg{\GYTVOS}{\GYTVOSH} & \hqg{\JYTunseen}{\JYTunseenH} & \hqg{\FYTunseen}{\FYTunseenH} & \hqg{\JYTseen}{\JYTseenH} & \hqg{\FYTseen}{\FYTseenH}
    }
}

\usepackage[pagebackref,breaklinks,colorlinks]{hyperref}

\usepackage[capitalize]{cleveref}
\crefname{section}{Sec.}{Secs.}
\Crefname{section}{Section}{Sections}
\Crefname{table}{Table}{Tables}
\crefname{table}{Tab.}{Tabs.}

\def\cvprPaperID{1615} 
\def\confName{CVPR}
\def\confYear{2022}

\begin{document}

\makeatletter
\DeclareRobustCommand\onedot{\futurelet\@let@token\@onedot}
\def\@onedot{\ifx\@let@token.\else.\null\fi\xspace}


\newcolumntype{Y}{>{\centering\arraybackslash}X}
\newcolumntype{Z}{>{\raggedleft\arraybackslash}X}

\newcommand{\ourMethodName}{HODOR\xspace}


\title{\ourMethodName: High-level Object Descriptors for Object Re-segmentation\\ in Video Learned from Static Images}

\author{
Ali Athar$^1$ %
\quad
Jonathon Luiten$^{1,2}$
\quad
Alexander Hermans$^1$%
\quad
Deva Ramanan$^2$%
\quad
Bastian Leibe$^1$\\[5pt]
$^1$RWTH Aachen University, Germany \quad $^2$Carnegie Mellon University, USA \\[5pt]
{\tt\small \{athar,luiten,hermans,leibe\}@vision.rwth-aachen.de}
\quad {\tt\small deva@cs.cmu.edu} \\
}


\maketitle

\begin{abstract}

Existing state-of-the-art methods for Video Object Segmentation (VOS) learn low-level pixel-to-pixel correspondences between frames to propagate object masks across video. This requires a large amount of densely annotated video data, which is costly to annotate, and largely redundant since frames within a video are highly correlated.
In light of this, we propose \ourMethodName: a novel method that tackles VOS by effectively leveraging annotated static images for understanding object appearance and scene context.
We encode object instances and scene information from an image frame into robust high-level descriptors which can then be used to re-segment those objects in different frames. 
As a result, \ourMethodName achieves state-of-the-art performance on the DAVIS and YouTube-VOS benchmarks compared to existing methods trained without video annotations. 
Without any architectural modification, \ourMethodName can also learn from video context around single annotated video frames by utilizing cyclic consistency, whereas other methods rely on dense, temporally consistent annotations.
Source code is available at: \url{https://github.com/Ali2500/HODOR}

\end{abstract}

\section{Introduction}
\label{sec:intro}

Current state-of-the-art Video Object Segmentation (VOS) methods learn `space-time correspondences' (STC), \ie pixel-to-pixel correspondences, between the image frames in a video.
These methods~\cite{Oh19ICCV,Yang20ECCV,Cheng21NeurIPS} achieve impressive results, but require a large amount of temporally dense annotated video for training. Such datasets require significant human effort, and the annotations are largely redundant since image frames within a video are highly correlated. The largest publicly available VOS dataset~\cite{Xu18Arxiv} contains only a few thousand videos. Single image datasets~\cite{Lin14ECCV,Kuznetsova18Arxiv}, in contrast, exist with hundreds of thousands of annotated images.
In this work, we explore the following question: can VOS be learned with only single-image annotations?

\begin{figure}[t]
  \begin{subfigure}[b]{\linewidth}
    \includegraphics[width=\linewidth]{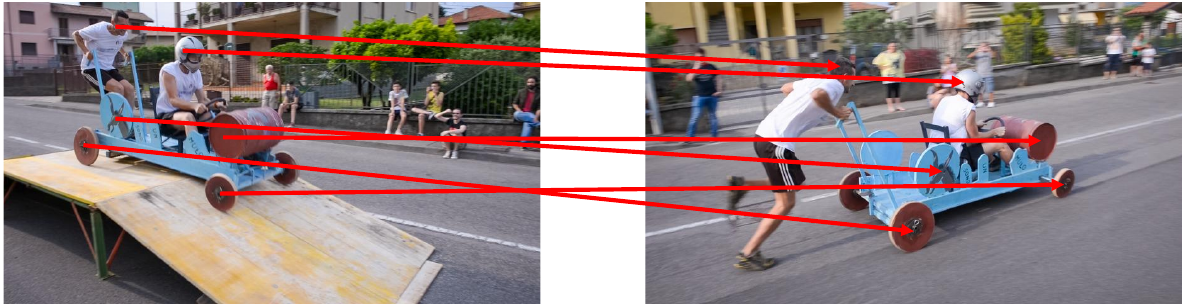}
    \caption{Space-time correspondence \cite{Oh19ICCV,Yang20ECCV,Cheng21NeurIPS,Seong20ECCV, Yang21NeurIPS,Cheng21CVPR,Seong21ICCV,Liang20NeurIPS,Lu20ECCV,Voigtlaender19CVPR}. \vspace{-9pt}}
    \label{fig:intro_old_methods}
  \end{subfigure}
  \noindent\rule{\linewidth}{0.5pt}
  \vspace{3pt}
  \begin{subfigure}[b]{\linewidth}
    \includegraphics[width=\linewidth]{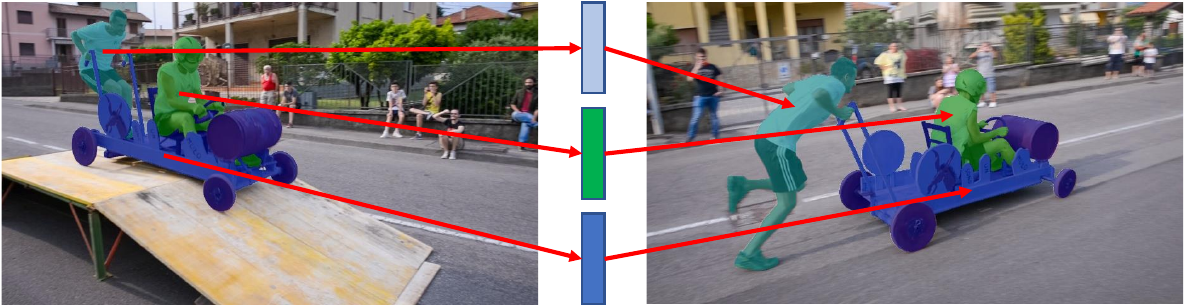}
    \caption{High-level Object Descriptors for Object Re-segmentation (ours).}
    \label{fig:intro_our_method}
  \end{subfigure}
  \caption{Previous methods (a) learn low-level pixel-pixel correspondence to propagate object masks. \ourMethodName (b) learns high-level object descriptors to re-segment objects in a different frame.}
  
  \label{fig:intro}
\end{figure}

\begin{figure*}[t]
  \centering
  \includegraphics[width=\linewidth]{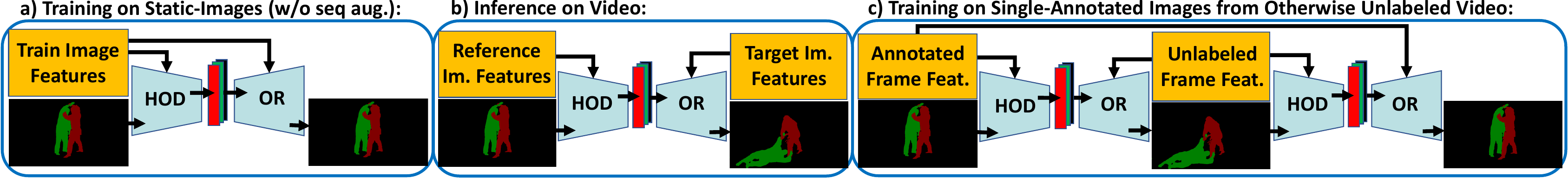}
  \caption{\textbf{\ourMethodName train and inference strategies.} HOD: High-level Object Descriptor Encoder. OR: Object Re-segmentation Decoder. Left: \ourMethodName can be trained with single annotated images (without sequence augmentations). Center: \ourMethodName is run on video by feeding features from a different frame to the decoder. Right: Training \ourMethodName can take advantage of unlabeled frames using cycle-consistency.}
  \label{fig:strategies}
\end{figure*}

%
To this end, we propose \ourMethodName: \textbf{H}igh-level \textbf{O}bject \textbf{D}escriptors for \textbf{O}bject \textbf{R}e-segmentation, a novel VOS framework which extracts a robust, \textit{high-level descriptor} for the given objects and background in an image. 
These descriptors are then used to find and segment those objects in another video frame, \ie \textit{re-segment} them, even if the object moves or changes appearance (Fig.~\ref{fig:intro_our_method}).
This differs fundamentally from STC methods which learn low-level, pixel-to-pixel correspondences (Fig.~\ref{fig:intro_old_methods}). 
%
%
The underlying idea is that high-level object descriptors can be learned without sequential video data, as this only requires understanding object appearance, and not reasoning about motion. Thus, \ourMethodName can be trained for VOS \textit{using only single images without any video motion augmentation} (Fig~\ref{fig:strategies}\textcolor{red}{a}), and still be applied to video (Fig~\ref{fig:strategies}\textcolor{red}{b}). This is inherently not possible with STC methods since learning correspondences requires comparing multiple, different frames.
%

%
The key to our approach is that it forces object appearance information to pass through a concise descriptor, \ie an information bottleneck. This prevents the descriptor from trivially summarizing the object mask shape and location.
%
The network thus learns to concisely encode object appearance, and also to match the descriptor to each pixel in order to re-segment the object in the same image.
%
%

If we add sequential augmentation to our single image training strategy to increase the network's robustness, \ourMethodName out-performs all existing methods trained with similar augmented image sequences on the DAVIS~\cite{Pont-Tuset17Arxiv} and YouTube-VOS~\cite{Xu18Arxiv} benchmarks.
This is because STC methods can only learn correspondences of simple motion from augmented frames, and thus cannot generalize well to the complex motion of real video. \ourMethodName however, being based on high-level object appearance and scene context, is much more resilient to this discrepancy. 



\ourMethodName can also be trained using cycle consistency on video where only a single frame is annotated (Fig~\ref{fig:strategies}\textcolor{red}{c}). Without modifying the approach at all, we can simply propagate masks through unlabeled frames and then in reverse back to the labeled frame to apply the loss. This is enabled by a fully differentiable formulation for attending to soft input masks which allows gradients to flow through multiple frame predictions. 
Based on this, our network can learn to be more robust to appearance changes that occur in natural video, while only requiring single annotated frames.
Current STC methods cannot be trained under this setting.

There are two further advantages:
%
%
The encoder can process, and model interactions between, an arbitrary number of objects. This improves performance and makes the inference speed largely independent of the number of objects. 
This is in contrast to many works~\cite{Oh19ICCV,Yang20ECCV,Cheng21CVPR,Cheng21NeurIPS} where part of the network requires separate forward passes per object. 
(2) The decoder can jointly attend to object descriptors over multiple past frames with negligible overhead. Thus, we can incorporate temporal history during inference even though the method can be trained on just single images.

To summarize: we propose a novel VOS framework that uses high-level descriptors to propagate objects across video. This enables training using just single images, with or without other unlabeled video frames. Our model processes an arbitrary number of objects simultaneously, and can readily incorporate temporal context during inference. We achieve state-of-the-art results on DAVIS and YouTube-VOS among methods trained without video annotation.

\section{Related Work}
\label{sec:related_work}

We group existing VOS methods into three categories: pixel-pixel, object-object and object-pixel. Though not all methods perfectly fit this taxonomy, it is nonetheless useful in comparing our approach to existing works.


\PAR{Pixel-pixel Correspondence.} Such approaches learn low-level space-time correspondence between pixels, and use these correspondences to propagate object masks between video frames. Whereas early VOS approaches~\cite{perazzi2017learning,Khoreva19IJCV,li2018video,cheng2017segflow} used pre-computed optical flow as a measure for pixel-pixel correspondence, FEELVOS~\cite{Voigtlaender19CVPR} was the first to learn these correspondences in an end-to-end fashion within the VOS framework, and STM~\cite{Oh19ICCV} significantly improved upon this. Nearly all subsequent VOS methods~\cite{Yang20ECCV,Cheng21NeurIPS,Seong20ECCV, Yang21NeurIPS,Cheng21CVPR,Seong21ICCV,Liang20NeurIPS,Lu20ECCV}, including the two current state-of-the-arts (STCN~\cite{Cheng21NeurIPS} and AOT-L~\cite{Yang21NeurIPS}) are based on the space-time correspondence paradigm, with each proposing various novel techniques for improving speed and performance.
\ourMethodName diverges from this paradigm by instead learning correspondences between pixels and high-level object descriptors.
%

\PAR{Self-supervised Pixel-pixel Correspondence.} 
One set of methods learns pixel-pixel correspondences using unlabelled video via self-supervision. To do this, some methods~\cite{Vondrick18ECCV,Lai20CVPR} optimize their network with colorization and image reconstruction based training objectives. Other methods~\cite{Wang19CVPR,Jabri20NeurIPS} learn from cyclic consistency by propagating random image patches through a video sequence. 
\ourMethodName can also be trained with cyclic consistency, but with the objective of learning high-level object descriptors rather than low-level pixel correspondences.
%

\PAR{Object-object Comparison.} Another common VOS approach involves directly comparing object representations~\cite{li2018video,Luiten18ACCV,zeng2019dmm,Voigtlaender20CVPR,Liang21ICCV}. Such methods first learn object proposals for the target image, and then match these proposals to previously tracked objects. 
%
%
This paradigm is inspired by methods in multi-object tracking~\cite{Voigtlaender19bCVPR,Bergmann19ICCV,wojke2017simple}, and often involves spatial similarity constraints and object ReID vectors~\cite{Hermans17Arxiv,li2018video} for temporal association. Such methods require training for proposal generation on a specific set of object classes, and thus do not generalize well to novel categories.
%

\begin{figure*}[t]
  \centering
\includegraphics[width=\linewidth]{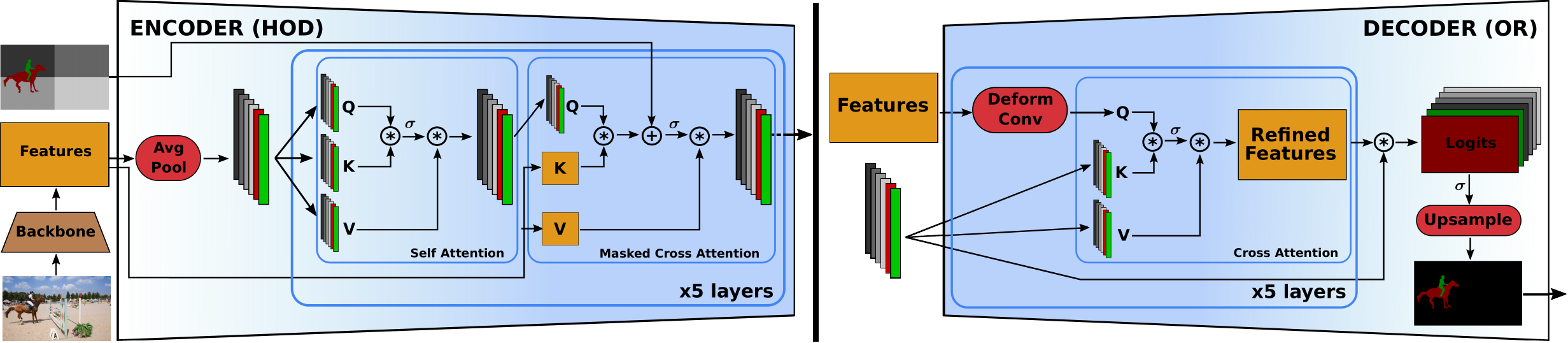}
  \caption{\textbf{The \ourMethodName Architecture} consists of a backbone, the HOD encoder, and the OR decoder. Q, K, and V refer to Queries, Keys and Values, respectively. The encoder jointly encodes all objects and background cells (here $2\times 2$) to descriptors, which are then decoded to masks by the decoder. Some steps are simplified  (the final upsampling) or omitted (fc layers, skip connections). See Sec.~\ref{sec:method} for details.}
  \label{fig:details}
\end{figure*}

\PAR{Object-pixel Comparison.} An alternative to the pixel-pixel and object-object approaches described above is learning high-level representations for the set of given objects, and then re-segmenting these objects in the target frame by directly comparing the representations to the pixel features in that frame. 
%
%
Early VOS methods followed this paradigm by finetuning a segmentation network during inference on the given first frame object masks~\cite{Caelles17CVPR,Voigtlaender17BMVC,Maninis18TPAMI,perazzi2017learning} to embed an object representation in the weights of a network, which is then applied directly to subsequent frames.
This is extremely slow and usually achieves poor results.
%
%
The most similar work to ours is SiamMask~\cite{wang2019fast}. It learns a vector representation for each object which is directly compared to pixel features to determine whether or not the pixels belong to that object. However, this approach trains on large amounts of annotated video data, and compared to \ourMethodName and other existing methods, does not achieve good results.
To the best of our knowledge, no other method from this category even achieves competitive results for VOS. 

After online finetuning based methods fell out of fashion, leaderboards for VOS benchmarks were dominated by object-object association based methods~\cite{li2018video,Luiten18ACCV} until the emergence of FEELVOS~\cite{Voigtlaender19CVPR} and STM~\cite{Oh19ICCV}. Since then, state-of-the-art VOS approaches are almost exclusively based on the pixel-pixel correspondence paradigm.

\section{Method}
\label{sec:method}

The \ourMethodName network architecture consists of three components: (1) a backbone which learns multi-scale image features, (2) a High-level Object Descriptor (HOD) encoder, and (3) an Object Re-segmentation (OR) decoder. With \ourMethodName, we revisit the idea of learning object-level descriptors for VOS which have mostly been replaced in favour of STC-based approaches. 
To this end, our network architecture enables the essence of an object to be encoded without directly memorizing the object mask's shape or location.
We also introduce attention layers which allow multiple objects to be processed simultaneously, and allow interactions between their descriptors. These attention layers also enable the descriptors to be enriched with image features (in the encoder), and vice versa (in the decoder).



The architecture is illustrated in Fig.~\ref{fig:details}. Given an RGB image $I \in \real{H\times W\times 3}$, the backbone produces a pair of $C$-dimensional feature maps $\mathcal{F} = \{ F^4, F^8 \}$ at the $4\times$ and $8\times$ downsampled input resolution scales, respectively.
\ext{Markus suggested to pull the next paragraph here, I kind of like the idea.}
Assume that the image $I$ contains $O$ objects of interest with segmentation masks $\mathcal{M}^f = \{ M^f_1, ..., M^f_O \}$.
We first compute a background mask consisting of all the pixels which do not belong to any object. This background mask is then split into $B$ separate masks $\mathcal{M}^b = \{ M^b_1, ..., M^b_B \}$ by dividing it into a grid with $B$ cells.

\subsection{Encoder}
\label{subsec:encoder}
%

The encoder accepts as input the set of masks $\mathcal{M}^f \cup \mathcal{M}^b$ and the image feature map $F^8$, and produces a set of object descriptors $\mathcal{D}^f = \{ \embedding{1}{f}, ... , \embedding{O}{f} \}$ containing one $C$-dimensional descriptor per foreground object, and likewise a set of descriptors $\mathcal{D}^b = \{ \embedding{1}{b}, ..., \embedding{B}{b} \}$ containing one $C$-dimensional descriptor for each background patch. Intuitively, these descriptors are a concise latent representation for their respective patches (object or background).
    
Each descriptor is initialized by average pooling the set of pixel features belonging to the corresponding patch. These are then iteratively and jointly refined by a series of transformer-like layers. 
%
Each layer consists of multi-head self-attention between the set of descriptors $\mathcal{D}^f \cup \mathcal{D}^b$, followed by multi-head cross-attention in which these descriptors absorb patch-specific information from the feature maps $F^8$ conditioned on the masks $\mathcal{M}^b \cup \mathcal{M}^f$. 
%
    
With some abuse of notation, let us use $D^{(l)} = \mathcal{D}^f \cup \mathcal{D}^b \in \real{(O+B) \times C}$ to denote the set of descriptors at the $l$-th layer of the encoder and $M = \mathcal{M}^f \cup \mathcal{M}^b \in \real{(O+B) \times H\times W}$ for the set of patch masks. The $l$-th layer of our encoder can then be described as follows:%
\begin{equation}
\label{eq:encoder}
\begin{split}
    D^{(l)} & \longleftarrow D^{(l-1)} +\smalltt{SelfAttn}(D^{(l-1)}) \\
    D^{(l)} & \longleftarrow D^{(l)} + \smalltt{MaskedCrossAttn}(D^{(l)}, F^8, M) \\
    D^{(l)} & \longleftarrow D^{(l)} + \smalltt{FFN}(D^{(l)})
\end{split}%
\end{equation}%
We omit the typical LayerNorm for the sake of text clarity (\cf~\cite{Carion20ECCV}). \smalltt{FFN} denotes a Feed-forward Network consisting of three fully-connected layers with ReLU activations. \smalltt{SelfAttn} denotes multi-head attention~\cite{Vaswani17NIPS} wherein queries, keys, and values are produced by applying separate linear projections to the input tensor.
\smalltt{MaskedCrossAttn} denotes multi-head attention where queries are produced from the embeddings, but keys and values are produced from the image feature map $F^8$. We describe this operation as `\smalltt{Masked}' because we condition the pixel features $F^8$ on the mask $M$ to enable the descriptors to better focus on their respective patches.
%
This could be done by replacing the dot product affinities between the descriptors $D$ and $F^8$ in the $\text{Key}^T \text{Query}$ matrix ($K^TQ$) with $-\infty$ for pixels where the respective mask value for the given patch is zero. However, this operation is not differentiable, and it restricts each descriptor to attend to only those pixels for which the corresponding patch mask value is one. 
Furthermore, numerical issues can arise for empty masks during training.

\PAR{Differentiable Soft Attention-masking.}
We propose a better formulation which is differentiable, allows the mask $M$ to be non-binary (\ie have soft values), and affords the network more flexibility to focus on relevant image features.
Given the mask $M \in [0,1]$ and a learnable, positive scalar $\alpha$, we define the attention operation as follows:%
%
\begin{equation}
    \label{eq:encoder_attn_masking}
    \smalltt{softmax} \left( \frac{K^T Q ~{\color{better_blue}+~ \alpha M}} {\sqrt{C}} \right) \cdot V
\end{equation}
This is identical to the standard attention operation proposed by Vaswani~\etal~\cite{Vaswani17NIPS} except for the term `{\color{better_blue}$+ \alpha M$}'. 
In practice, each attention head is assigned a different learnable parameter $\alpha$ which is optimized during training. Thus, different attention heads attend to pixel features conditioned on different magnitudes of masking. This enables the network to learn descriptors which focus on their respective patch features, but that are also able to capture scene information from other parts of the image if this is beneficial for the training objective. This is inspired by Press~\etal~\cite{Press21Arxiv} who used additive offsets in temporal attention in NLP.

The encoder is thus designed to learn descriptors conditioned on object/background masks. This formulation contains an inherent information bottleneck which does not allow the input mask's shape or location from directly `leaking' into the descriptors. Specifically, in Eq.~\ref{eq:encoder_attn_masking}, the mask $M$ can only influence the $\smalltt{softmax}(\cdot)$ term, \ie the weights with which the Values ($V$) are summed, but $M$ cannot directly be copied into the attention operation output.

\subsection{Decoder}
\label{subsec:decoder}

Whereas the encoder produces descriptors $\mathcal{D}^f \cup \mathcal{D}^b$ by conditioning the image features $\mathcal{F}$ on patch masks $\mathcal{M}^f \cup \mathcal{M}^b$, the decoder does the opposite: it (re)produces the patch masks $\mathcal{M}^f \cup \mathcal{M}^b$ by conditioning the image features $\mathcal{F}$ on the descriptors $\mathcal{D}^f \cup \mathcal{D}^b$.
The architecture is similar to that of the encoder consisting of a series of transformer-like layers with multi-head attention. However, now the image feature map $F^8$ is updated iteratively by attending to the descriptors. There are two additional differences: (1) the cross-attention does not involve any masking. (2) The self-attention cannot be used for feature maps with large spatial dimensions due to its quadratic memory complexity, so we instead use a $3\times 3$ deformable convolution~\cite{Dai17ICCV}.
Since the purpose of self-attention is to enrich pixel features by allowing them to attend to all other pixels, a deformable convolution can be thought of as having a similar effect where a pixel can interact with a set of other pixels at learned offsets.
Although recent works~\cite{Zhu20ArxivDeformableDetr,Dai21ICCVDynamicDetr} proposed efficient variants of attention for image features, we found that deformable convolutions still require less memory. 

Let us denote with $F^{8(l)}$ the feature map at the $l$-th layer of the decoder, and let us use $D \in \real{(O+B)\times C}$ to denote the descriptors produced by the encoder. The $l$-th decoder layer can then be described as (again omitting LayerNorms):%
\begin{equation}
\label{eq:decoder_1}
\begin{split}
    F^{8(l)} & \longleftarrow F^{8(l-1)} + \smalltt{DeformConv}(F^{8(l-1)}) \\
    F^{8(l)} & \longleftarrow F^{8(l)} + \smalltt{CrossAttn}(F^{8(l)}, D) \\
    F^{8(l)} & \longleftarrow F^{8(l)} + \smalltt{FFN}(F^{8(l)})
\end{split}
\end{equation}
For the \smalltt{CrossAttn}, a linear projection generates the queries from the feature map $F^{8(l)}$ and the keys and values are two separate linear projections of the descriptors $D$.
We omit the final \smalltt{FFN} from Fig.~\ref{fig:details} for space reasons.

The final decoder layer outputs a feature map $F^{8(L)}$, which we bilinearly upsample by a factor of 2 and then add to the image feature map $F^4 \in \mathcal{F}$.
We then apply a $3\times 3$ convolution to get $F^{4(L)}$ and at this scale we compute the per-pixel object logits based on the dot product between $F^{4(L)}$ and the descriptors $D$.
The resulting logits are upsampled to the input resolution, before applying a \smalltt{softmax} over the descriptor dimension, yielding the output masks $M$. Formally, $M \in \real{H\times W\times (O+B)}$ is calculated as follows:%
%
%
%
\begin{equation}
\label{eq:decoder_2}
\begin{split}
    F^{4(L)} & \longleftarrow \smalltt{Conv} \left( F^4 + \smalltt{upsample2}(F^{8(L)}) \right)\\
    M & \longleftarrow \smalltt{softmax} \left( \smalltt{upsample4}(F^{4(L)} \cdot D) \right) \\
\end{split}
\end{equation}

\subsection{Video Object Segmentation}
\label{subsec:vos}

So far we discussed how the decoder can reproduce the patch masks which were input to the encoder. However, since the  descriptors encode a robust representation for the objects in an image, the decoder can re-segment them in any image $I'$ where these objects exist. 
Let us use $I_t$, $\mathcal{F}_t$, $\mathcal{D}_t$ and $\mathcal{M}_t$ to denote the image frame, feature maps, descriptors, and masks at frame $t$ of a given video clip, respectively. Given the first frame $I_1$ of a $T$-frame clip, and the segmentation masks $\mathcal{M}^f_1$ for $O$ objects in the first frame, we can learn a set of descriptors $\mathcal{D}_1^f \cup \mathcal{D}_1^b$ which encode these objects as well as the background (\cf Sec.~\ref{subsec:encoder}). We can then segment these objects in another video frame $I_t$ by simply giving the decoder the feature maps $\mathcal{F}_t$ for that frame and conditioning it on the first-frame descriptors $\mathcal{D}_1^f \cup \mathcal{D}_1^b$. 

This strategy, however, would not generalize well to lengthy videos with significant scene changes and where objects intersect and occlude each other. In practice, we therefore propagate the object masks sequentially frame-by-frame: $t: 1 \xrightarrow{} 2 \xrightarrow{} ... \xrightarrow{} T$. At each frame $t$, the encoder creates an updated set of object descriptors $\mathcal{D}_t^f \cup \mathcal{D}_t^b$ from the masks $\mathcal{M}_{t-1}$ predicted for the previous frame (or the initial input masks when $t-1=1$). The decoder then segments the objects in frame $t$ using these updated descriptors.

\PAR{Temporal History.} 
To mitigate large object appearance and scene changes in video, existing VOS methods~\cite{Oh19ICCV,Yang20ECCV,Cheng21NeurIPS} incorporate temporal context from multiple past frames when predicting object masks for the current frame. \ourMethodName also achieves the same feat efficiently: recall that the decoder is conditioned on the set $\mathcal{D}^f \cup \mathcal{D}^b$ which contains a variable number of object/background descriptors. 
To incorporate temporal history when predicting masks for frame $t$, we simply take the union of the set of descriptors for the past $T_p$ frames which we want to incorporate, \ie $\mathcal{D}_{t-T_p}^f \cup \mathcal{D}_{t-T_p}^b \cup ...  \cup \mathcal{D}_{t-1}^f \cup \mathcal{D}_{t-1}^b$.
%
In the decoder, feature map $F^8_t$ will be refined by jointly attending to the set of all descriptors in the $T_p$ frame history. The subsequent dot-product with the descriptors will produce a set of masks $M_t \in \real{H\times W\times T_p\times (O+B)}$. We temporally aggregate over the time dimension to obtain masks for each of the $O+B$ patches (we use $\smalltt{max}$ for our method).

This formulation has three advantages: 
%
(1) it incurs little computational overhead since we only need the $O+B$ descriptors for each past frame instead of the full feature maps. 
(2) We can train with only single images, and still incorporate temporal context during inference without any architectural changes.
(3) We can segment an arbitrary number of objects with a single forward pass of the network. This is in contrast to several VOS methods (\eg \cite{Oh19ICCV,Cheng21NeurIPS}) which require a per-object forward pass for at least part of the network.
\todo{add more citations for multi forward pass vos methids.}

\subsection{Training}
\label{subsec:training}

%
\ourMethodName's problem formulation makes it versatile with respect to the type of training data it can utilize. For the basic setting, we only need a static image dataset with annotated object masks. However, annotated image sequences, if available, can also be utilized by simply propagating the object (and background) masks over the given sequence.

Furthermore, our problem formulation enables the sequential propagation of object masks over a video to be end-to-end differentiable, \ie even if we only supervise the masks predicted for the last frame of a given clip, the error will be backpropagated over the entire temporal sequence to the first frame. This allows \ourMethodName to also be trained on unlabeled frames
from videos with arbitrarily sparse and temporally inconsistent object ID annotations. Given a training clip with $T$ frames where only frame $t=1$ is annotated, we can propagate the given object masks from $t:1 \xrightarrow{} T$, and then further propagate them in reverse temporal order from $t:T \xrightarrow{} 1$. We can then use the principle of cyclic consistency~\cite{Jabri20NeurIPS,Wang19CVPR} for supervision by supervising the predicted masks for $t=1$ to be identical to the input masks.
The inherent information bottleneck of our method enables it to be trained effectively under this setting without trivially copying the input masks across the sequence. 


\begin{figure*}[t]
  \centering
  \setlength\tabcolsep{0pt}
  \renewcommand{\arraystretch}{0.5}
  \begin{tabularx}{\textwidth}{YYYcY}%
       \includegraphics[width=0.184\textwidth]{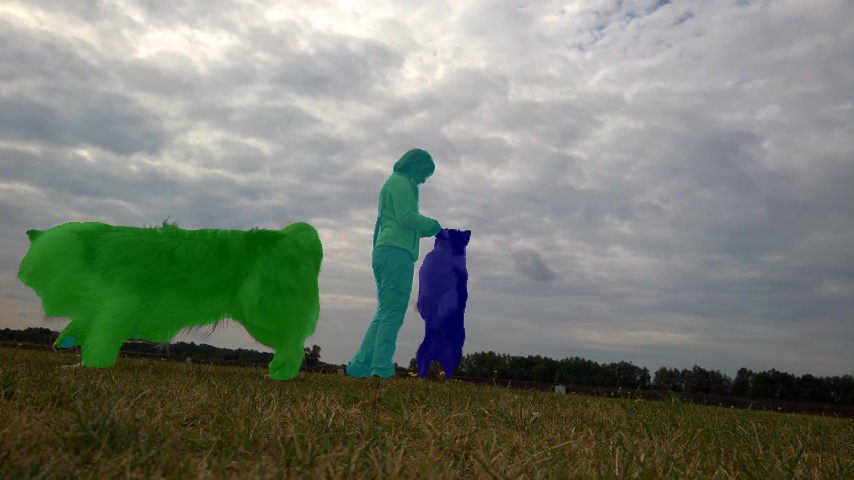}&%
       \includegraphics[width=0.184\textwidth]{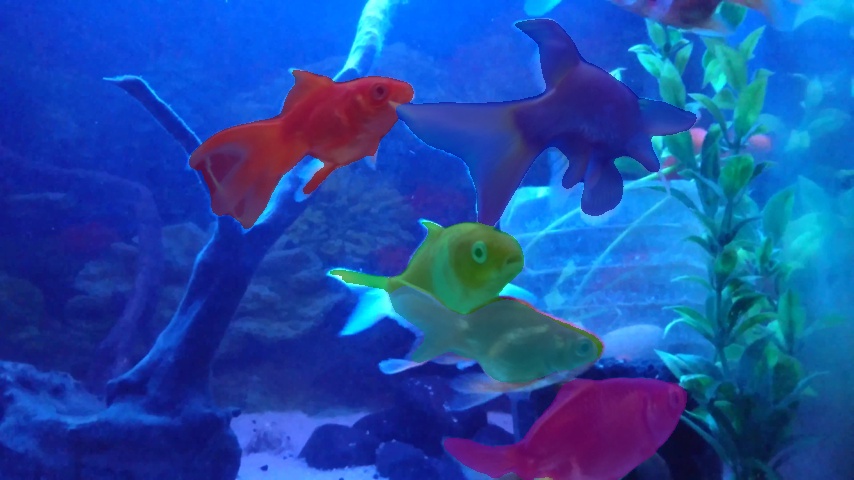}&%
       \includegraphics[width=0.184\textwidth]{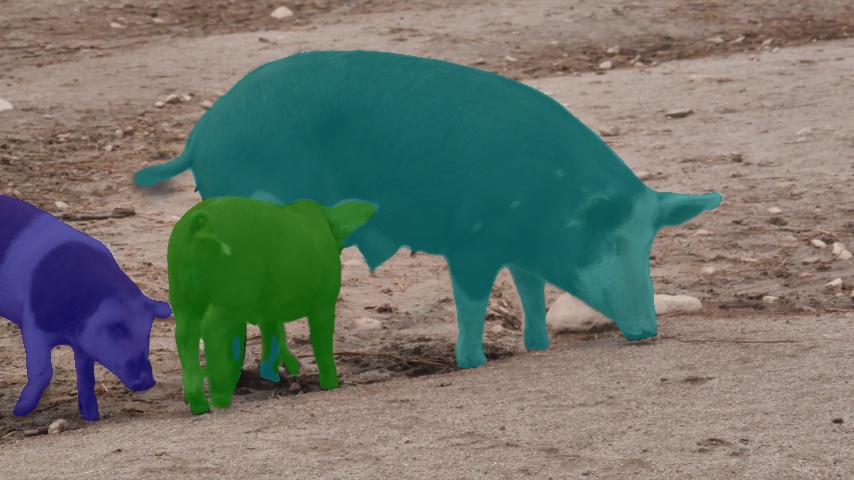}&%
       \includegraphics[width=0.248\textwidth]{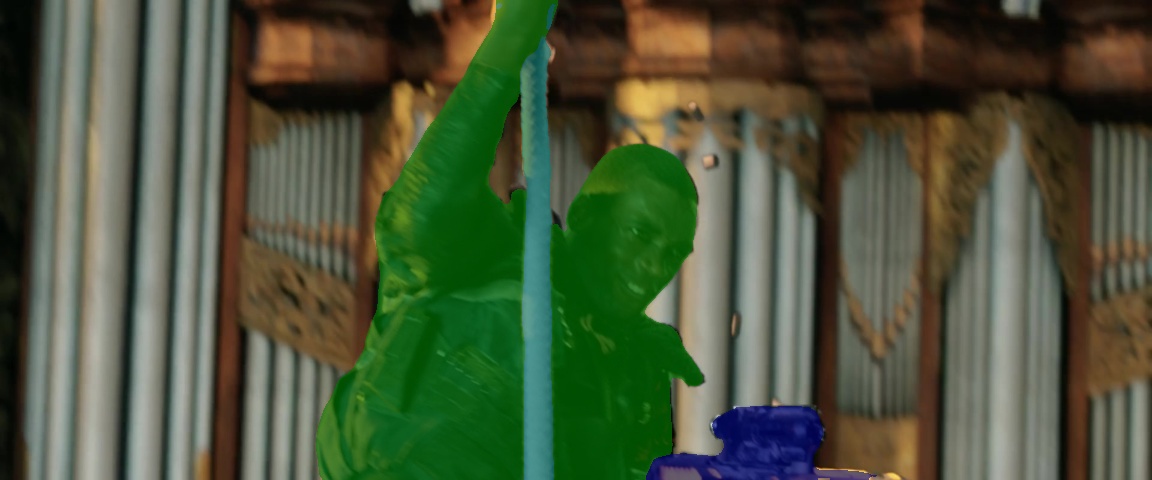}&%
       \includegraphics[width=0.184\textwidth]{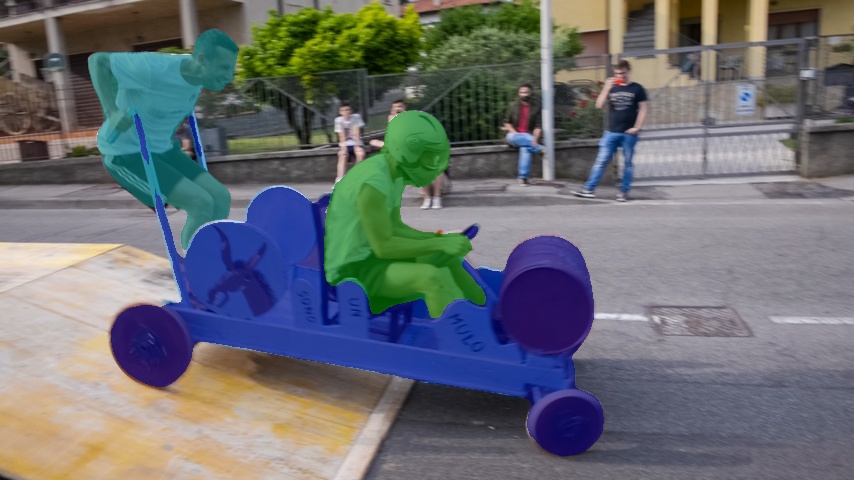}\\%
       \includegraphics[width=0.184\textwidth]{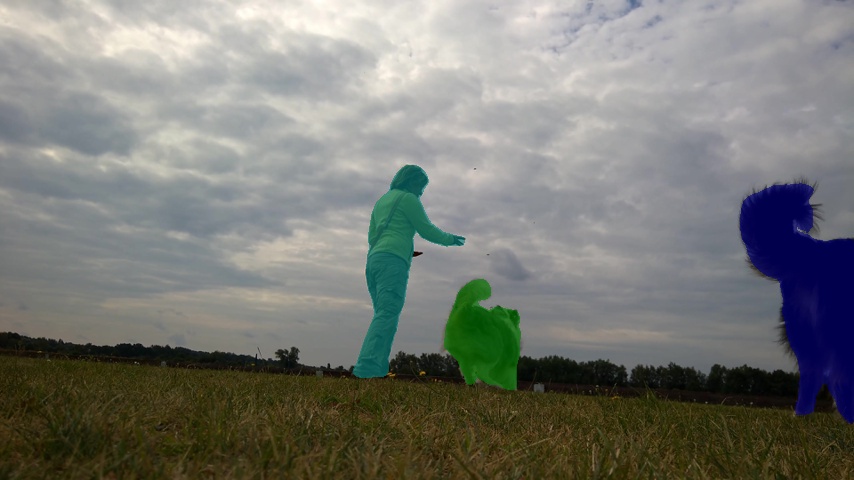}&%
       \includegraphics[width=0.184\textwidth]{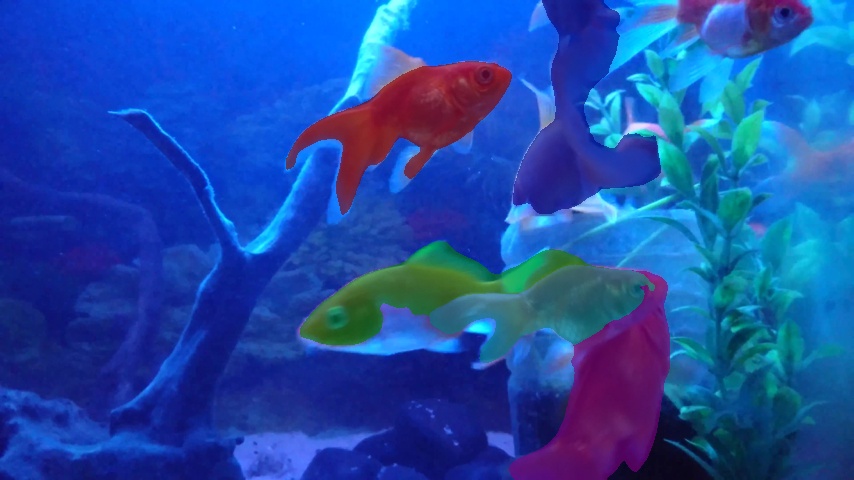}&%
       \includegraphics[width=0.184\textwidth]{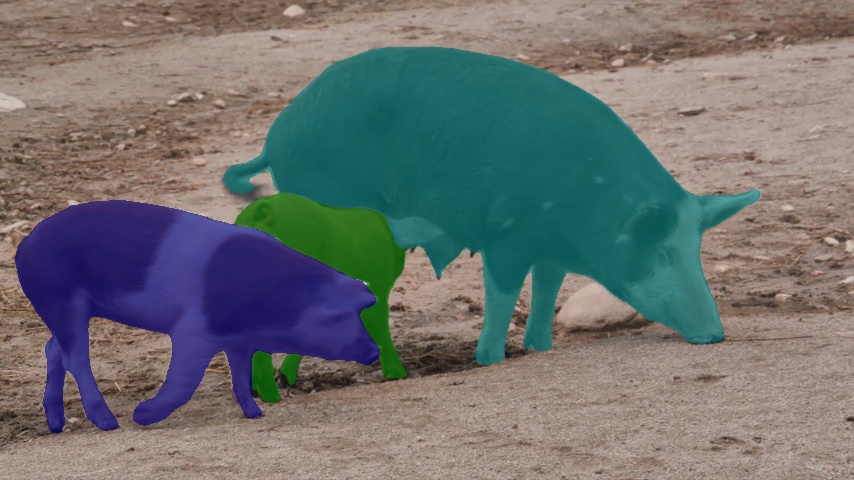}&%
       \includegraphics[width=0.248\textwidth]{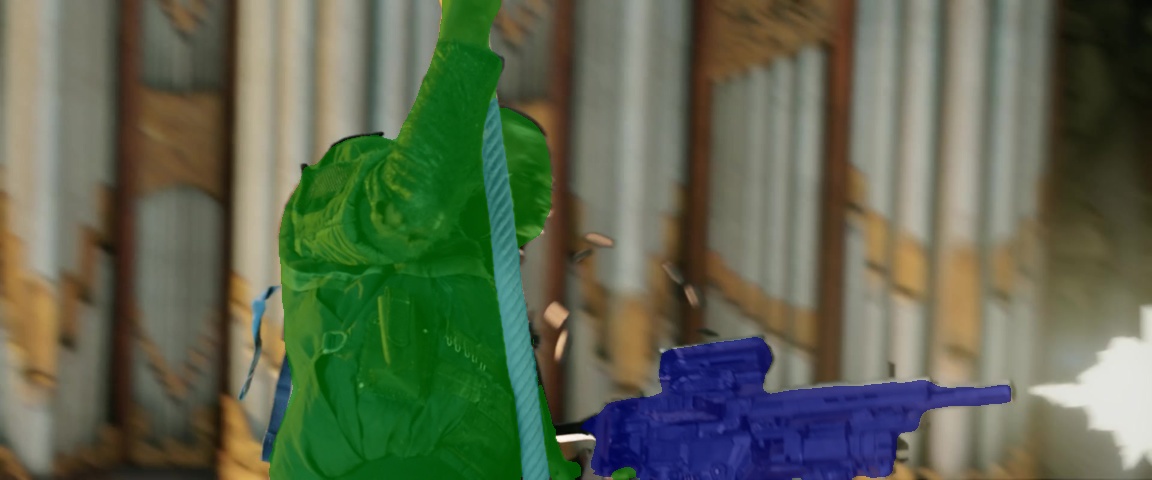}&%
       \includegraphics[width=0.184\textwidth]{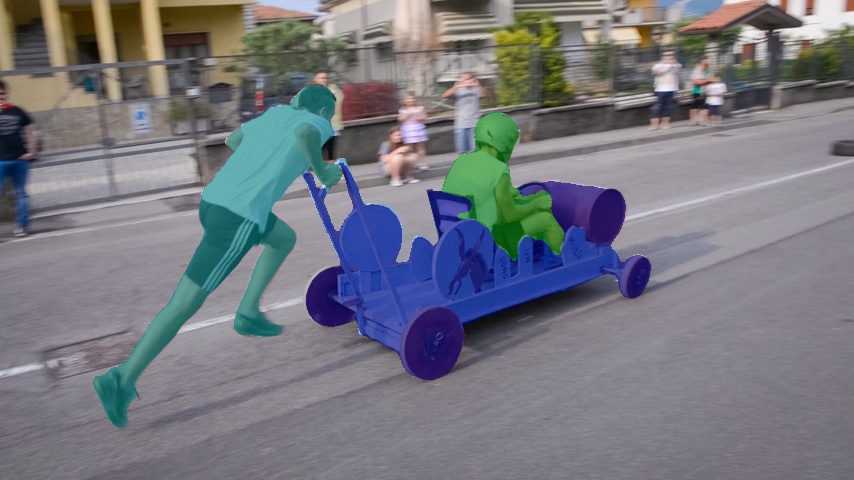}\\%
  \end{tabularx}
  \caption{\textbf{Qualitative results on the DAVIS 2017 validation set:} We omit the given first frame masks and only show results for other frames. Note that fish, pigs, rope, guns, and soapbox carts are not annotated in COCO, which is used to train our model.}
  \label{fig:visual_results}
\end{figure*}

\section{Experimental Evaluation}
\label{sec:experiments}

\PAR{Datasets.} We evaluate \ourMethodName on the DAVIS'17~\cite{Pont-Tuset17Arxiv} and YouTube-VOS 2019~\cite{Xu18Arxiv} benchmarks. The DAVIS dataset comprises 60, 30, and 30 video sequences for training, validation and testing, respectively. YouTube-VOS is a larger dataset with 3471 videos for training and 507 for validation. For both benchmarks the task is to segment and track an arbitrary number of objects in each video. The ground truth mask for each object is only provided for the first frame in which an object appears. The evaluation measures are the $\J$ score (Jaccard Index), $\F$ score ($\text{F}_1$-score) and the average of the two ($\JnF$) is treated as the final measure. 

\newcommand{\scoreCocoDavisVal}{77.5\xspace}

\PAR{Implementation Details.}
Our backbone network is the `Tiny' variant of the Swin transformer~\cite{Liu21Arxiv} with Feature Pyramid Network (FPN) and both our encoder and decoder consist of 5 layers. 
For all training settings involving static images, we use the COCO~\cite{Lin14ECCV} dataset. Whenever training on image sequences, each sequence contains $T=3$ frames.
We provide between 1 and 4 randomly chosen labeled objects per image/sequence. 
The encoder and decoder weights are randomly initialized, whereas the backbone is initialized from an off-the-shelf checkpoint trained for object detection~\cite{Lin14ECCV}. The model is trained using the AdamW optimizer~\cite{Loshchilov17Arxiv} with a batch size of 8 parallelized across 4 Nvidia RTX3090 GPUs. 
During inference, we use a temporal history comprising 7 past frames. The inference runs at $\sim\hspace{-2pt}17$ frames/s on an Nvidia RTX3090, independent of the number of instances.
See supp. material for further implementation details \eg learning rate schedule, training time.


\subsection{Training Data Versatility}
\label{subsec:training_data_types}

Table~\ref{tab:data_modalities} shows results on the DAVIS'17 validation set for \ourMethodName trained under different settings. For comparison, we also report results for STCN~\cite{Cheng21NeurIPS}, the current state-of-the-art VOS method, whenever applicable.
%
%
On just single images, \ourMethodName achieves 61.6 $\JnF$, which is at the level of state-of-the-art VOS approaches from 2017~\cite{Caelles17CVPR,Voigtlaender17BMVC} that use online fine-tuning during inference. STCN inherently requires an image sequence and cannot be trained in this setting. 
In row 2, we train on image sequences generated by duplicating the same image $T$ times without any augmentation. Whereas existing space-time correspondence based methods collapse under this setting by learning to trivially copy the input mask, \ourMethodName achieves 69.4 $\JnF$. Even though this setting does not provide any extra `information' to the network compared to row 1, the $\JnF$ increases significantly from 61.6 to 69.4. 
This is because the model experiences noisy input masks due to the sequential propagation involved in this training setting. Thus, during inference the model can robustly track objects across lengthy videos even if the intermediate frame masks are imprecise, as it has encountered similar masks during training.
%
%

In row 3, we train on image sequences generated by applying $T$ random affine transformations to static images. With this setting, our $\JnF$ further improves to 77.5. This is because such augmentations coarsely approximate video motion, thus making the learned object descriptors more robust to object appearance and scene changes. Existing VOS methods also ubiquitously train on such augmented image sequences as a pretraining step, however we out-perform them under this setting (77.5 vs. 75.8).
Qualitative results of this model can be seen in Fig.~\ref{fig:visual_results} and an analysis of the object descriptors is given in the supplementary material.

\begin{table}[t]
\centering{}

\small



\caption{$\JnF$ scores for various training settings on the DAVIS 2017 validation set. The sequence length $T=3$ in all experiments. CC: Cyclic Consistency.
}

\label{tab:data_modalities}

\begin{tabularx}{\linewidth}{clYY}
\toprule 
 & Training Setting & \ourMethodName & STCN~\cite{Cheng21NeurIPS} \tabularnewline
\midrule
1 & Single image & 61.6 & -\tabularnewline
\arrayrulecolor{lightgray}\midrule[0.25pt]\arrayrulecolor{black}
2 & $T\times$ duplicated image (no aug) & 69.4 & -\tabularnewline
\arrayrulecolor{lightgray}\midrule[0.25pt]\arrayrulecolor{black}
3 & $T\times$ duplicated image (with aug) & 77.5 & 75.8\tabularnewline
\midrule
4 & $T\times$ dupl. video frame (with aug) & 79.0  & 72.8 \tabularnewline
\arrayrulecolor{lightgray}\midrule[0.25pt]\arrayrulecolor{black}
5 & $T$ frames, 1 annotated (with CC) & 80.6 & -\tabularnewline
\midrule
6 & Temporally dense video & 81.3 & 85.4\tabularnewline
\bottomrule

\end{tabularx}

\end{table}

We then explore how effectively \ourMethodName can leverage single frame annotations that are part of a video sequence.
For the next two experiments, we utilize the YouTube-VOS~\cite{Xu18Arxiv} and DAVIS~\cite{Pont-Tuset17Arxiv} training sets, but assume that only one frame per video (the middle-most frame) is annotated (we only use 3,531 of the 98,797 available video frame annotations). In row 4, we fine-tune models from row 3 by similarly augmenting the selected frames. This further improves the $\JnF$ from 77.5 to 79.0. STCN on the other hand performs worse (72.8), likely because of overfitting.

For row 5, we fine-tune the model from row 3, but this time using cyclic consistency by randomly sampling $T-1$ unlabeled frames around the single annotated frame in each video. This improves the $\JnF$ from 77.5 to 80.6. The fact that this is higher than the 79.0 $\JnF$ in row 4 shows that \ourMethodName can effectively learn video motion cues from unlabeled frames. Existing supervised STC methods cannot be trained with this strategy since they lack the information bottleneck needed to prevent the network from trivially copying the input mask and also because they cannot back-propagate gradients through the predicted mask.

\begin{table*}[t]
    \centering
        \caption{Quantitative results on the DAVIS and YouTube-VOS datasets.
        For YouTube-VOS we focus on the 2019 validation set, but substitute 2018 validation set results when only those are available (slightly higher, highlighted in 
        \textcolor{gray}{grey}). As is common, we evaluate unseen (us) and seen (s) object classes separatly for Youtube-VOS, UI$^{\dagger}$: Unlabeled Images, OL: Online Fine-tuning, $^{\ast}$: retrained by us.}
    \label{tab:main_results}
    
\small
    
\begin{tabularx}{\linewidth}{p{0.3cm}lcYYYcYYYcYYYYY}
\toprule
&&& \multicolumn{3}{c}{DAVIS val 17} && \multicolumn{3}{c}{DAVIS test-dev 17} && \multicolumn{5}{c}{YouTube-VOS \textcolor{gray}{val 18} /val 19}\\
\cmidrule{4-6} \cmidrule{8-10} \cmidrule{12-16}
&  & OL & $\mathcal{J} \& \mathcal{F}$ & $\mathcal{J}$ & $\mathcal{F}$  && $\mathcal{J} \& \mathcal{F}$ & $\mathcal{J}$ & $\mathcal{F}$  && $\mathcal{J} \& \mathcal{F}$ & $\mathcal{J}_{us}$ & $\mathcal{F}_{us}$ & $\mathcal{J}_{s}$ & $\mathcal{F}_{s}$
\\
\midrule

UI$^{\dagger}$%
\tableline{DINO}{DINO \cite{Caron21ICCV}}
\midrule

\multirow{11}{*}{\rotatebox{90}{Labeled images}}%
\tableline{OSVOS}{OSVOS \cite{Caelles17CVPR}}
\tableline{OnAVOS}{OnAVOS \cite{Voigtlaender17BMVC}}
\tableline{OSVOS_S}{OSVOS\textsuperscript{S} \cite{Maninis18TPAMI}}
\tablelineold{STM_im_only}{STM (5x Mix) \cite{Oh19ICCV}}
\tableline{DMN_AOA_COCO}{DMN+AOA (COCO) \cite{Liang21ICCV}}
\tableline{KMN_im_only}{KMN (5x Mix) \cite{Seong20ECCV}}
\tableline{STCN_im_only}{STCN (5x Mix) \cite{Cheng21NeurIPS}}
\tableline{CFBI_COCO}{CFBI (COCO) \cite{Yang20ECCV} $^{\ast}$}
\tableline{STCN_COCO}{STCN (COCO) \cite{Cheng21NeurIPS} $^{\ast}$}
\tableline{Ours_COCO_aug}{\textcolor{better_blue}{\ourMethodName (Ours, COCO)}}
\tableline{Ours_COCO_CC}{\textcolor{better_blue}{\ourMethodName (Ours, COCO + CC)}}


%

\midrule

\multirow{3}{*}{\footnotesize \rotatebox{90}{Unlabeled} \rotatebox{90}{\hspace{4pt} videos}}%
%
\tableline{MAST}{MAST \cite{Lai20CVPR}}
\tableline{STC-CRW}{STC-CRW  \cite{Jabri20NeurIPS}}
\tableline{MAMP}{MAMP~\cite{Miao21Arxiv} (uses optical flow)} 

\midrule

\multirow{11}{*}{\rotatebox{90}{Labeled videos}}%
%
\tableline{FEELVOS}{FEELVOS \cite{Voigtlaender19CVPR}}
\tablelineold{AFB-URR}{AFB-URR \cite{Liang20NeurIPS}}
\tablelineold{e-OSVOS}{e-OSVOS \cite{Meinhardt20NeurIPS}} 
\tablelineold{STM}{STM \cite{Oh19ICCV}}
\tableline{CFBI}{CFBI \cite{Yang20ECCV}}
\tableline{EG-VOS}{EG-VOS \cite{Lu20ECCV}} 
\tablelineold{KMN}{KMN \cite{Seong20ECCV}}
\tablelineold{DMN_AOA}{DMN+AOA \cite{Liang21ICCV}}
\tableline{HMMN}{HMMN \cite{Seong21ICCV}}
\tableline{STCN}{STCN \cite{Cheng21NeurIPS}}
\tableline{AOT-L}{AOT-L \cite{Yang21NeurIPS}}

\bottomrule

\end{tabularx}
\end{table*}

Finally, we also train using dense video annotations with full supervision (row 6), which improves the $\JnF$ from 77.5 to 81.3. STCN out-performs us under this setting (85.4 $\JnF$) because the same information bottleneck which enables us to train on single images and unlabeled frames with cyclic consistency also has the drawback of limiting the network's access to fine-grained video motion cues. By contrast, pixel-to-pixel correspondence methods lack such a bottleneck thus enabling them to better leverage dense video data. Nonetheless, to the best of our knowledge, we are the first to surpass 81 $\JnF$ on DAVIS'17 using an approach not based on pixel-to-pixel correspondences. Note also that our approach has much better scaling properties since we require only one frame annotation per video as opposed to the dense annotations required by existing methods.

\subsection{Comparison to State-of-the-art}
\label{subsec:main_results}



In Table~\ref{tab:main_results}, we report results for existing VOS methods categorized by the type of training data used. Results for \ourMethodName are given for two settings: (1) when trained on augmented image sequences from COCO~\cite{Lin14ECCV} (\cf Table~\ref{tab:data_modalities}, row 3), and (2) after fine-tuning with cyclic consistency using just one labeled frame per training set video (\cf Table~\ref{tab:data_modalities}, row 5).
%
We use the same model checkpoint for all three benchmarks.
%
For the sake of completeness we also list results for methods that do not require any annotations, and also those trained on densely annotated video. 

Looking at the `Labeled Images' category, it can be seen that \ourMethodName trained on COCO achieves 77.5 $\JnF$ on DAVIS'17, outperforming all existing methods. This includes earlier VOS methods~\cite{perazzi2017learning,Caelles17CVPR,Voigtlaender17BMVC,Maninis18TPAMI} that perform online fine-tuning (best score: 68.0 $\JnF$ from $\text{OSVOS}^\text{S}$~\cite{Maninis18TPAMI}), but also current state-of-the-art methods which pre-train on similar augmented image sequences. The best performing method among these is STCN (75.8 $\JnF$) which is 1.7 $\JnF$ lower than our 77.5. It is worth noting that while DMN+AOA~\cite{Liang21ICCV} use COCO images for this training step, STM~\cite{Oh19ICCV}, KMN~\cite{Seong20ECCV} and STCN~\cite{Cheng21NeurIPS} use a collection of 5 image datasets~\cite{Wang17CVPRImgDS1,Shi15PAMIImgDS2,Zeng19ICCVImgDS3,Cheng20CVPRImgDS4,Li20CVPRImgDS5} (`5x Mix' in the table).
To verify that this discrepancy does not disadvantage other methods, we retrained STCN and CFBI on COCO images using their respective training code. These experiments are marked with `$\ast$' in the table. It can be seen that STCN performs significantly worse on DAVIS under this setting (55.0 $\JnF$). Though performance on YouTube-VOS is comparatively better (69.4 $\JnF$), \ourMethodName still outperforms it (71.7 $\JnF$). The same trend holds true for CFBI. 
One possible explanation for the large performance difference of these methods on the two datasets is that the augmentations applied to static images are quite aggressive, which make objects undergo significant movement across frames. This better approximates YouTube-VOS videos where objects also frequently undergo large motions. By contrast, object motion in videos from DAVIS is comparatively milder, and because these methods learn pixel-to-pixel correspondences, they do not perform well during inference if the nature of object motion is different from what was encountered during training. 

Finally, we also report our result after fine-tuning on cyclic consistency using only the middle-most annotated frame from each video in the YouTube-VOS and DAVIS training set. This improves the $\JnF$ by 3.1, 1.0 and 0.7 points on the DAVIS validation, DAVIS test and YouTube-VOS validation sets, respectively.

\subsection{Ablations}
\label{subsec:ablations}

We perform ablations to investigate our design choices and report the results in Table~\ref{tab:ablations}.

\begin{figure}[t]
  \centering
  \includegraphics[width=0.9\linewidth]{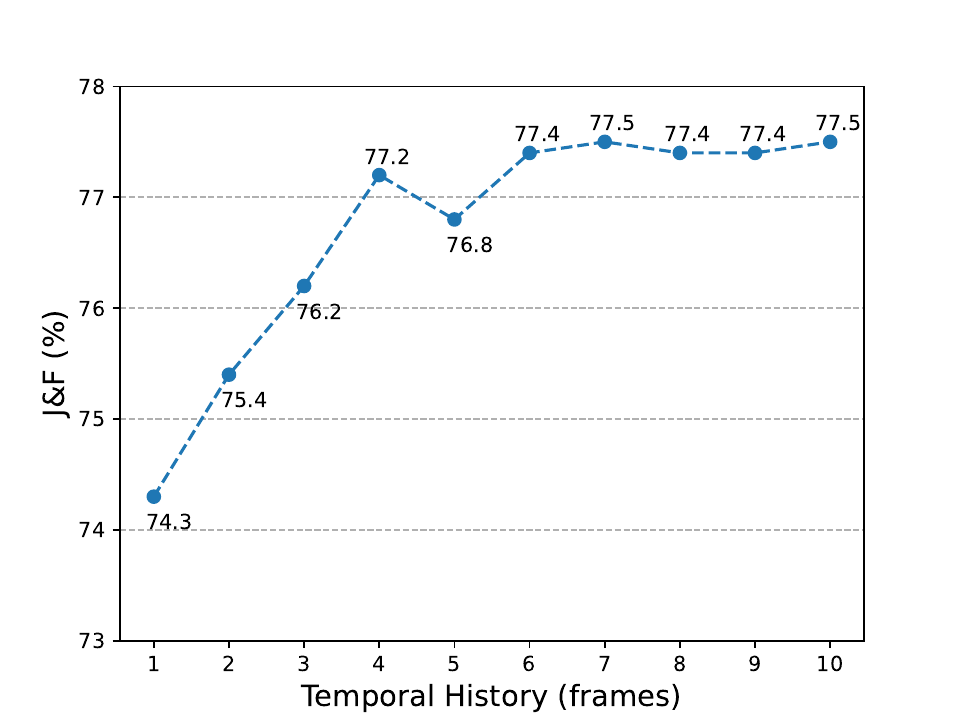}
  \caption{\textbf{Temporal History Ablation:} Performance on DAVIS'17 val for different temporal history lengths during inference.}
  \label{fig:inference_temporal_history}
\end{figure}

\PAR{Temporal History During Inference.}
In Sec.~\ref{subsec:vos}, we discussed how HODOR can effectively incorporate temporal history from past frames when predicting the object masks for a given frame. Fig.~\ref{fig:inference_temporal_history} plots the $\JnF$ score on DAVIS'17 val for different temporal history lengths. It can be seen that increasing the frame history from 1 to 4 frames yields an approximately linear performance improvement from 74.3 to 77.2. Thereafter, the $\JnF$ saturated at 7 frames at a $\JnF$ score of 77.5. Recall that because we only need the object/background descriptors for past frames rather than the full feature maps, the inference run-time is minimally affected by the temporal history length: increasing the temporal history from 1 to 10 frames only reduces the inference speed from 17.3 to 16.7 frame/s (reported speed is an average over 5 runs).

\PAR{Multi-instance.} Unlike most other VOS methods, \ourMethodName handles all objects in a single forward pass.
This enables it to effectively utilize multi-object context and learn better descriptors.
For experiment (1), we train and infer with only one foreground instance, and merge multiple forward passes by running a pixel-wise \smalltt{argmax} over the object logits during inference.
This reduces the $\JnF$ from \scoreCocoDavisVal to 71.5, clearly highlighting the benefit of our multi-instance approach which also increases inference speed by negating the need for per-object forward passes.
%

\PAR{Attention Masking in Encoder.} Recall from Sec.~\ref{subsec:encoder} that we condition the descriptors on their respective patch masks with our proposed soft attention mechanism. Experiment (2) shows the result without attention masking. Thus, the only cue for the descriptors to specialize to their respective targets is their initialization (average pooling over the target pixel features); this reduces the $\JnF$ to 74.4. For experiment (3), we apply hard attention masking by thresholding the masks at 0.5 and setting the $\text{K}^T\text{Q}$ matrix entries inside the attention operation to $-\infty$ for pixels where the mask is zero. This strategy yields a similarly reduced $\JnF$ of 74.5.
The performance increase from using our learned soft attention masking shows that it helps the encoder to better condition the descriptors on the given object/background masks.

\PAR{Encoder/Decoder Layers.} For experiments (4-7), we ablate the depth of the encoder and decoder.
Reducing the depth from 5 to 3 in either of them reduces the $\JnF$ by $\sim\hspace{-2pt}1$.
For the zero-layer case, the 5 encoder layers are replaced by a single MLP consisting of 3 fully-connected layers and the 5 decoder layers are replaced with two $3\times 3$ convolutions.
%
%
%
For the zero-layer decoder, the $\JnF$ reduces to 74.4, whereas the zero-layer encoder reduces the $\JnF$ to 72.8.
This shows that while both components play an important role in the overall performance, the encoder has a larger impact.
One reason could be that whereas self-attention in the encoder allows object descriptors to interact, such interactions are not as profound in the decoder where deformable convolutions are used instead of self-attention.

\PAR{Barebones Network.} 
For experiment (8), we completely omit both encoder and decoder, \ie the descriptors are generated by simply average pooling the backbone features, and the output masks are generated by computing their dot product with backbone features of a different image. Doing so reduces the $\JnF$ from 77.5 to 70.6. This shows that although the method does not completely collapse without the encoder/decoder layers, the latter are still important and impart an improvement of 6.9 $\JnF$.

\PAR{Deformable Convolution in Decoder.}
Due to memory constraints, we use a $3\times 3$ deformable convolution~\cite{Dai17ICCV} instead of the self-attention operation (\cf Sec.~\ref{subsec:decoder}).
In experiment (9), we instead use a regular $3\times 3$ convolution and observe a reduction in $\JnF$ from \scoreCocoDavisVal to 75.1. This highlights the importance of substituting self attention with an operation that is able to attend to far-away spatial locations.

\PAR{Background Descriptors.} We use nine background descriptors  initialized by dividing the image into a $3\times 3$ grid and average pooling the background pixel features in each cell. This gives \ourMethodName more flexibility to model the background. For experiment (10) we instead use a single background descriptor, reducing the $\JnF$ from \scoreCocoDavisVal to 76.2.


\begin{table}[t]
\centering%
\caption{Several ablation results on the DAVIS 2017 validation set.}%
\label{tab:ablations}%
\small%
\begin{tabularx}{\linewidth}{clYYY}
\toprule 
 & Setting & $\mathcal{J\&F}$ & $\mathcal{J}$ & $\mathcal{F}$\tabularnewline
\midrule 
1 & Single foreground instance & 71.5 & 69.2 & 73.9 \tabularnewline
\arrayrulecolor{lightgray}\midrule[0.25pt]\arrayrulecolor{black}

2 & No masking in encoder & 74.4 & 71.5 & 77.2 \tabularnewline
3 & Hard masking in encoder & 74.5 & 71.8 & 77.1\tabularnewline
\arrayrulecolor{lightgray}\midrule[0.25pt]\arrayrulecolor{black}

4 & \# layers in encoder: $5 \xrightarrow{} 0$ & 72.8 & 70.5 & 75.2 \tabularnewline
5 & \# layers in encoder: $5 \xrightarrow{} 3$ & 76.6 & 73.9 & 79.4 \tabularnewline
\arrayrulecolor{lightgray}\midrule[0.25pt]\arrayrulecolor{black}

6 & \# layers in decoder: $5 \xrightarrow{} 0$ & 74.4 & 71.7 & 77.1 \tabularnewline
7 & \# layers in decoder: $5 \xrightarrow{} 3$ & 76.4 & 73.6 & 79.3 \tabularnewline
\arrayrulecolor{lightgray}\midrule[0.25pt]\arrayrulecolor{black}

8 & Barebones Network & 70.6 & 68.2 & 72.9 \tabularnewline
\arrayrulecolor{lightgray}\midrule[0.25pt]\arrayrulecolor{black}

9 & Regular convolution in decoder & 75.1 & 72.0 & 78.2 \tabularnewline
\arrayrulecolor{lightgray}\midrule[0.25pt]\arrayrulecolor{black}

10 & 1x background descriptor & 76.2 & 73.7 & 78.7 \tabularnewline

\midrule[0.25pt] 
 &  \ourMethodName & 77.5 & 74.7 & 80.2\tabularnewline
\bottomrule
\end{tabularx}
\end{table}

\section{Discussion}
\PAR{Limitations.}
Aside from our performance on dense video data (discussed in Sec.~\ref{subsec:training_data_types}), another limitation of our method is that when there are distractor objects in the scene with similar appearances, \ourMethodName sometimes compels itself to segment an object even if that object has moved out of the video scene. Since such cases arise more frequently in YouTube-VOS videos, this is one reason why our $\JnF$ score for YouTube-VOS is lower than that for DAVIS. In future work, improved training strategies could be formulated to better optimize the model for such challenging cases. 


\PAR{Ethical Considerations.}
As with most computer vision methods, the dual-use dilemma can and should not be ignored. However, it is unlikely that our approach could be utilized to facilitate negative use-cases (\eg population tracking or surveillance) more effectively than dedicated approaches for these applications.
%
Another important ethical aspect is that dataset annotation is often performed by an exploited labor force deprived of minimum wage and/or legally binding benefits. 
%
Reducing the need for such annotations can thus be seen as a positive aspect of our approach.

\section{Conclusion}

We proposed a novel VOS approach which uses high-level descriptors for encoding and propagating objects across video. Our approach contains an information bottleneck which enables training on single images and unlabeled frames using cyclic consistency. Thus, unlike existing STC based methods which train on dense video data, \ourMethodName can be trained on static images, or on videos with arbitrarily sparse, temporally inconsistent frame annotations. Since annotating single frames is easier than dense video, \ourMethodName has strong potential for scaling up performance by learning from large-scale video datasets with sparse, or even automatically generated frame annotations~\cite{Voigtlaender21WACV}.



\iftoggle{comments}{
{\color{red}
\vspace{20pt}
\textbf{Important deadline TODOs}
\begin{enumerate}
    \item define re-segment in the intro.
    \item Search for ??, [], etc. Also check warnings
\end{enumerate}
}
{\color{orange}
What we need/want for the supplementary:
\begin{enumerate}
    \item Detailed eval of embedding space, pca, t-sne, PR for global and sequence wise.
    \item Performance of used history plot
    \item Actually write about the catch all
    \item Visualization of what the background queries do (here also discuss the catch all one?)
    \item Visualization of stuff before the max operation which frames help more> t-3 or t-1?
    \item A visual investigation of training setups 1-3, why do repeated images work better than single images? Why does augmentaiton help?
    \item Visualize the different heads as Deva was asking for, could be cool to see if some heads focus on ``parts'' of objects
    \item Do the 0 enc 0 dec layer abblation? To search for the secret sauce needed to make our model work
    \begin{enumerate}
        \item Same but also no mlp and no 3x3 conv \\
        \item Maybe further explore what is import cross or self attention in encoder/ decoder?
    \end{enumerate}
    \item An actual hodor scene xD?
    \item Ethical Consideration
    \item implementation details
    \begin{itemize}
        \item resizing images train/test
        \item learning rate schedule
        \item catch all
        \item loss info
        \item mention the exact checkpoint.
    \end{itemize}
    \item upload some video?
\end{enumerate}
}
}{}

\PAR{Acknowledgements.}
This project was funded in part by ERC Consolidator Grant DeeVise (ERC-2017-COG-773161). Computing resources were granted by RWTH Aachen University under project `supp0003'. We thank Paul Voigtlaender, Jonas Schult, Istvan Sarandi, Sabarinath Mahadevan, Markus Knoche, Christian Schmidt, Jason Zhang, Gautam Gare and Tarasha Khurana for helpful discussions.

\bibliographystyle{abbrv}
\bibliography{abbrev_short,references}

\clearpage
\twocolumn[{%
 \centering
 \LARGE \textbf{Supplementary Material\\[1cm]}
}]

\setcounter{equation}{0}
\setcounter{figure}{0}
\setcounter{table}{0}
\setcounter{page}{1}
\setcounter{section}{0}
\makeatletter
\renewcommand{\theequation}{S\arabic{equation}}
\renewcommand{\thefigure}{S\arabic{figure}}
\renewcommand{\thetable}{S\arabic{table}}
\renewcommand{\thesection}{S\arabic{section}}

\section{Object Descriptors for Re-Identification}

\begin{figure}[h]
  \centering
  \includegraphics[width=\linewidth]{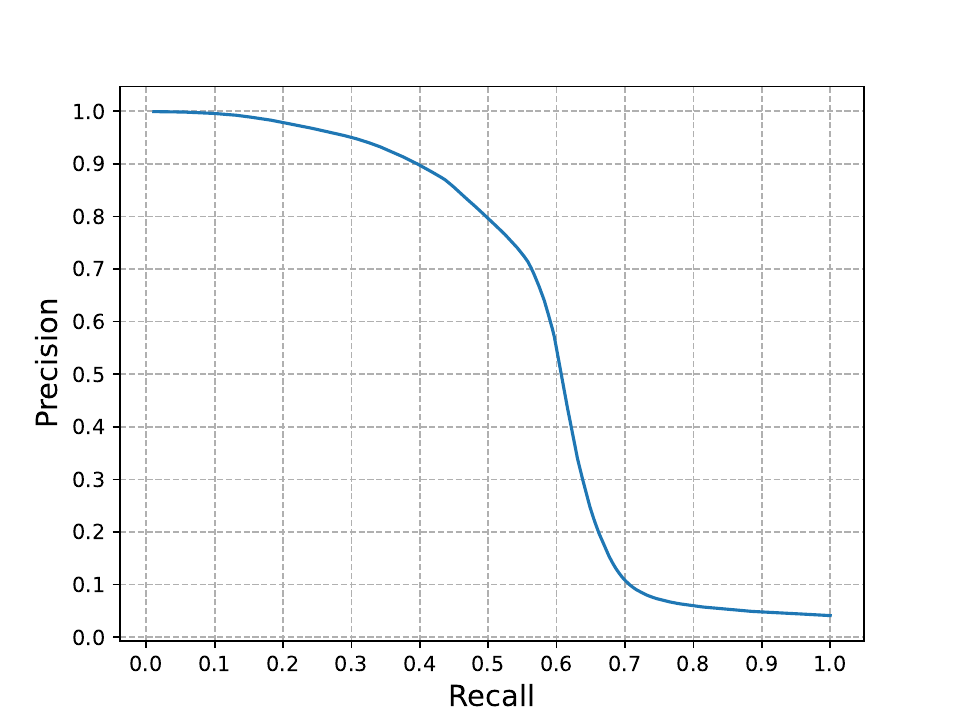}
  \caption{\textbf{Retrieval Task:} Precision-Recall curve for retrieval task on all object instances in DAVIS'17 val.}
  \label{fig:retrieval_pr_curve}
\end{figure}

Our object descriptors are trained to encode an object's appearance so that it can be re-segmented, \ie segmented in another video frame. Here we explore the applicability of these descriptors for a re-identification/retrieval task. For this, we consider the set of object descriptors for all frames for all video sequences in the DAVIS'17 validation set. For each descriptor, we calculate the Euclidean distance to all other descriptors, and then use these distances to retrieve other descriptors belonging to the same object instance. The resulting precision and recall is used to generate the precision-recall curve in Fig.~\ref{fig:retrieval_pr_curve} by averaging the retrieval scores across all descriptors.

Looking at the curve, we see that for each descriptor, $\sim$ 50\% of the other descriptors belonging to the same object instance can be retrieved with a fairly high recall of $\sim$ 80\%. Thereafter, the precision drops off sharply. Note, however, that this plot does not reflect the full quality of the object descriptors for the Video Object Segmentation (VOS) task due to two main reasons:

\begin{enumerate}

    \item This experiment disregards the image feature maps and directly compares the descriptors to one another. In the actual VOS use-case, we compute the dot-product between descriptors and image features to produce per-pixel logits which are then optimized to correctly segment the given object. In this experiment however, we directly compute the Euclidean distance between the descriptors themselves. Recall from Sec.~2 of the main text where we discussed that "Object-object Correspondence" based method use such re-identification techniques for associating objects over time. HODOR by contrast is an "Object-pixel Correspondence" based method.
    
    \item For this experiment, we expect the network to learn descriptors which separate objects globally, \ie across different video sequences. During training however, the network was only trained to distinguish between objects in the same image (or image sequence).
    
\end{enumerate}

We hence conclude from this experiment that the object descriptors learned by our network can be used for re-identification tasks. However, the distribution of the descriptors for a given instance do not follow a unimodal distribution. This results in the sharp drop-off in recall seen in Fig.~\ref{fig:retrieval_pr_curve}.

\section{Visualizing Descriptor Feature Space}

\begin{figure*}[t]
  \centering
  \includegraphics[width=\linewidth]{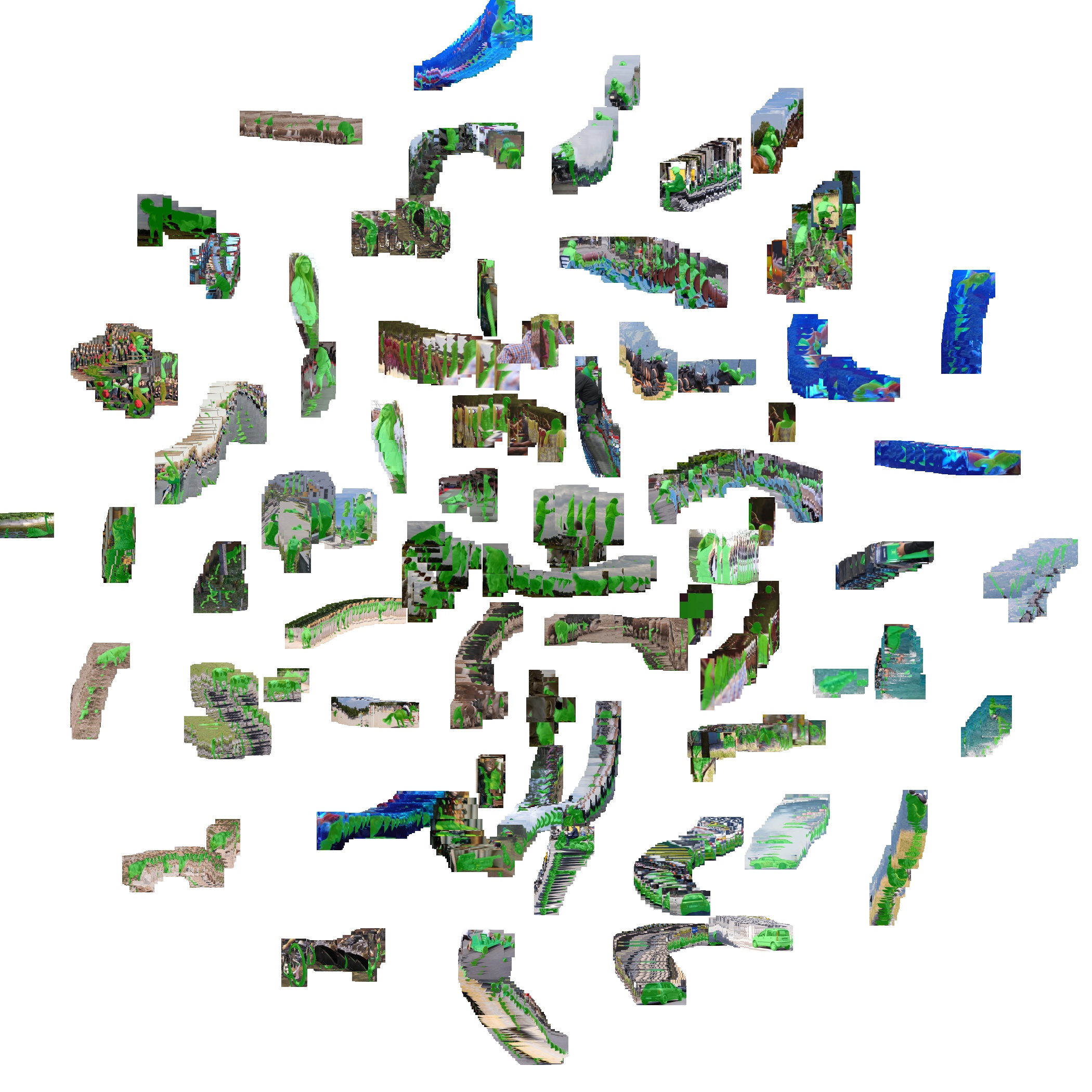}
  \caption{\textbf{Object Descriptor Visualization:} The descriptors for all object instances in DAVIS'17 val projected to 2-D using t-SNE.}
  \label{fig:descriptor_tsne_vis}
\end{figure*}

We attempt to visualize the object descriptors by projecting the 256-D object descriptors for all object instances in the DAVIS'17 validation to 2-D using t-SNE~\cite{Van08Tsne}. The resulting visualization is shown in Fig.~\ref{fig:descriptor_tsne_vis} wherein the object crop for each descriptor is pasted at the projected 2-D coordinates. We can clearly see that descriptors for the same object instance are tightly clustered in a trajectory-like sequence. Though not visualized here, we observed that the trajectory-like shape usually corresponds to the frame index, which means that the descriptors tend to drift slightly over time.

We can also see a strong semantic trend in the descriptors. The lower-right portion of the image contains several of the `\textit{car} objects, the top-right contains several \textit{riders} (\ie persons righting motorbikes, bicycles, horses). The center portions generally contains persons, and the lower-left portion of the image contains several animal classes \eg \textit{cow}, \textit{dog}, \textit{goat}. There are, however, noticeable exceptions. Note how there is a cluster of three fish on center-right, but the remaining two fish are very far away from them and each other. 

\begin{figure*}[t]
  \centering
  \includegraphics[width=\linewidth]{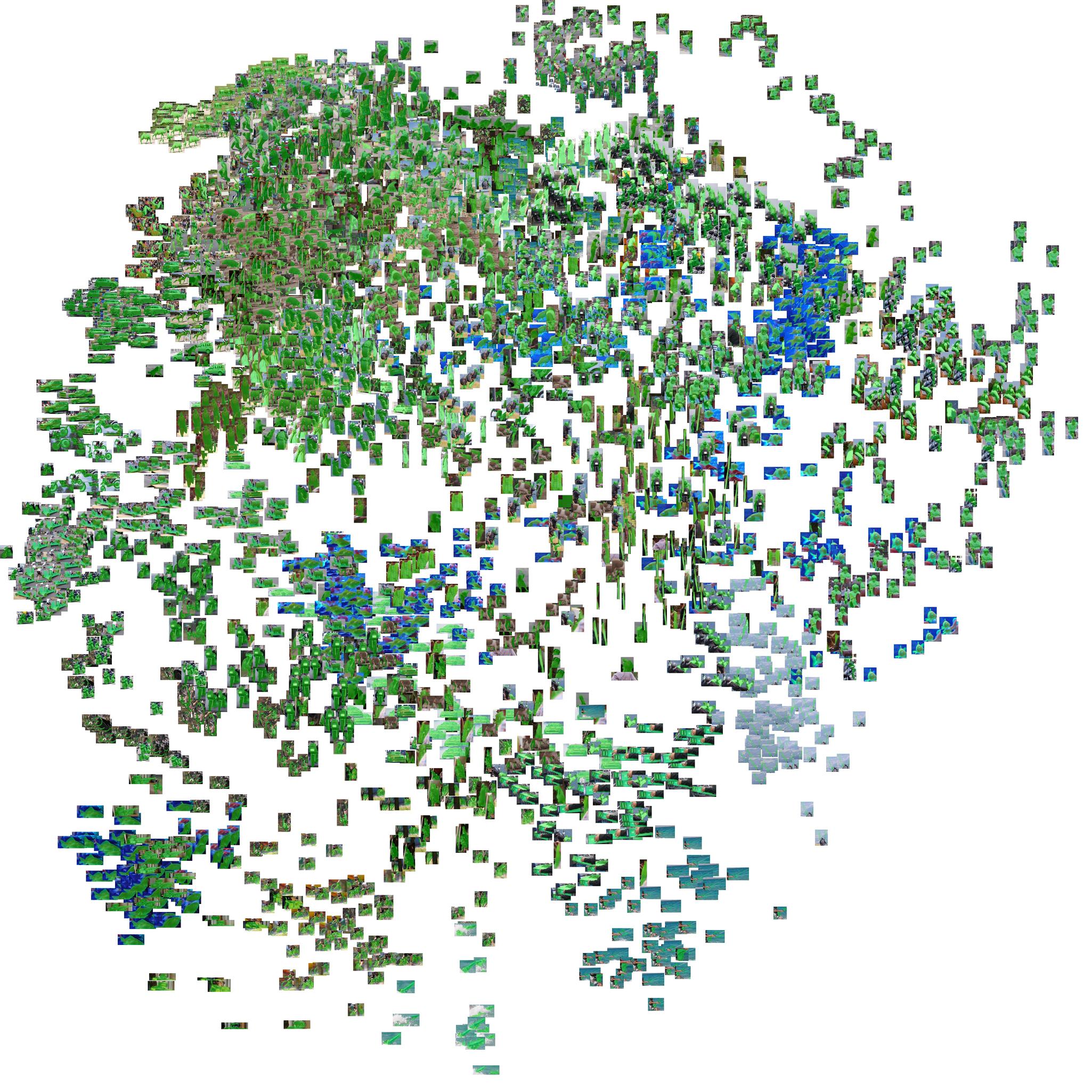}
  \caption{\textbf{Object Descriptor Visualization:} The descriptors for all object instances in DAVIS'17 val projected to 2-D using PCA.}
  \label{fig:descriptor_pca_vis}
\end{figure*}

Given that applications based on object embeddings often use a simple linear projection for further processing, we also visualize the object descriptors by projecting them to 2-D using Principal Component Analysis (PCA). The resulting illustration is given in Fig.~\ref{fig:descriptor_pca_vis}. Here, in general, the descriptors are less distinguishable from each other, but the overall trend still holds true, \ie descriptors for the same instance and similar semantic classes are generally located close to one another.

\section{Background Descriptors}

In Sec.~\ref{sec:method} of the main text, we explained how HODOR uses high-level descriptors to model the foreground objects and also the background. For the latter, all non-object pixels are combined into a background mask which is then split into 9 separate masks by dividing it into a $3\times 3$ grid. One further minor architectural detail 
is that aside from the 9 background descriptors, we also predict an additional `catch-all' background logit for each pixel. To do this, we apply a $3\times 3$ convolution followed by a $1\times 1$ convolution to the refined feature map $F^{4(L)}$ in the decoder (\cf Eq.~\ref{eq:decoder_2} in the main text) to obtain a single-channel logit map. Then, before computing the softmax over the descriptors, we append the logit value for each pixel, representing another background descriptor. Formally speaking, Eq.~\ref{eq:decoder_2} of the main text changes to the following:

\begin{equation}
\label{eq:supp_decoder_2}
\begin{split}
    F^{4(L)} & \longleftarrow \smalltt{Conv} \left( F^4 + \smalltt{upsample2}(F^{8(L)}) \right)\\
    M_{bc} & \longleftarrow \smalltt{Conv} \left( \smalltt{Conv} \left( F^{4(L)} \right) \right) \\
    M' & \longleftarrow \smalltt{Concatenate} \left( F^{4(L)} \cdot D, M_{bc} \right) \\
    M & \longleftarrow \smalltt{softmax} \left( \smalltt{upsample4} \left( M' \right) \right) \\
\end{split}
\end{equation}

where $M_{bc}$ is the background catch-all logit map and $M'$ is an intermediate variable used to denote the concatenation of the dot-products $F^{4(L)} \cdot D$ and $M_{bc}$.

Note that these catch-all background logits are not propagated frame-by-frame when processing a video sequence. Without this technique, we obtain a $\JnF$ of 76.2 on DAVIS'17 val, which is 1.3 lower than the 77.5 reported in Table~\ref{tab:main_results} of the main text. 

Some example probability heatmaps for both the catch-all, as well as the $3\times 3$ background grid can seen in Fig.~\ref{fig:background_desciptors} and \ref{fig:background_desciptors2}. Note how the catch-all logits have high magnitudes mostly around object edges. The background descriptors sometimes associate themselves to an object-like region \eg to the bush in the \textit{horsejump-high} sequence (bottom-right descriptor), or to the black box/case in the \textit{bike-packing} sequence (bottom-right descriptor). In general, we an also see a location bias based on the background mask patch which each descriptor was made to focus on by the encoder.

\newcommand{\bggrid}[2]{
    \setlength\tabcolsep{0pt}
    \renewcommand{\arraystretch}{0.5}
    \begin{tabularx}{\textwidth}{YYYY}%
      \includegraphics[width=0.243\textwidth]{images/supplementary/desc_vis/#1/#2.jpg}&%
      \includegraphics[width=0.243\textwidth]{images/supplementary/desc_vis/#1/#2_p0_i2.jpg}&%
      \includegraphics[width=0.243\textwidth]{images/supplementary/desc_vis/#1/#2_p0_i3.jpg}&%
      \includegraphics[width=0.243\textwidth]{images/supplementary/desc_vis/#1/#2_p0_i4.jpg}\\%
      \includegraphics[width=0.243\textwidth]{images/supplementary/desc_vis/#1/#2_p0_i1.jpg}&%
      \includegraphics[width=0.243\textwidth]{images/supplementary/desc_vis/#1/#2_p0_i5.jpg}&%
      \includegraphics[width=0.243\textwidth]{images/supplementary/desc_vis/#1/#2_p0_i6.jpg}&%
      \includegraphics[width=0.243\textwidth]{images/supplementary/desc_vis/#1/#2_p0_i7.jpg}\\%
      &%
      \includegraphics[width=0.243\textwidth]{images/supplementary/desc_vis/#1/#2_p0_i8.jpg}&%
      \includegraphics[width=0.243\textwidth]{images/supplementary/desc_vis/#1/#2_p0_i9.jpg}&%
      \includegraphics[width=0.243\textwidth]{images/supplementary/desc_vis/#1/#2_p0_i10.jpg}\\%
    \end{tabularx}
}
\newcommand{\bgviscapt}{Each block shows the ground truth foreground object mask(s) (top left) and a total of 10 background probability heatmaps, corresponding 1 catch-all heatmap (mid left) and the full $3 \times 3$ background grid heatmaps (columns 2-4). Not the location bias in the $3 \times 3$ grid, where the grid-based background descriptor initialization sometimes causes the background descriptors to attach to a nearby object.}
\begin{figure*}[t]
  \centering
  \bggrid{horsejump-high}{00005}
  \noindent\rule{\linewidth}{0.5pt}\vspace{2pt}
  \bggrid{bike-packing}{00000}
  \noindent\rule{\linewidth}{0.5pt}\vspace{2pt}
  \bggrid{dogs-jump}{00033}
  \caption{\textbf{Background Descriptor Visualization:} \bgviscapt}
  \label{fig:background_desciptors}
\end{figure*}

\begin{figure*}[t]
  \centering
  \bggrid{breakdance}{00002}
  \noindent\rule{\linewidth}{0.5pt}\vspace{2pt}
  \bggrid{soapbox}{00038}
  \noindent\rule{\linewidth}{0.5pt}\vspace{2pt}
  \bggrid{shooting}{00003}
  \caption{\textbf{Background Descriptor Visualization (continued):} \bgviscapt}
  \label{fig:background_desciptors2}
\end{figure*}

\section{Implementation Details}

\PAR{Input Image Dimensions.} For training, the input image is resized in an aspect-ratio preserving manner such that the pixel area is $\sim$ 300,000 and the lower dimension is an integer multiple of 32. During inference, the images are resized to have lower dimension 512.

\PAR{Loss Function.} To supervise the predicted masks, we use the sum of the cross-entropy loss and the DICE loss (both weighted by unity).

\PAR{Learning Rate Schedule.} When training on COCO~\cite{Lin14ECCV}, the learning rate is first warmed up from $0$ to $10^{-4}$ over 10k iterations. Then at 100k iterations we apply step decay and reduce the learning rate to $10^{-5}$. The training is then run for a further 150k iterations. For training on annotated video frames (both augmented frames and cyclic consistency), we fine-tune the network by loading weights from the COCO augmented sequence checkpoint, and then warm-up the learning rate from 0 to $10^{-5}$ over 10k iterations. The network then trains for a further 10k iterations with constant learning rate. Since there are only $\sim 3500$ labeled image frames under this setting, the model tends to over-fit if trained longer.

\PAR{Training Time.} The main training on COCO for 250k iterations requires $\sim 2$ days on 4 Nvidia 3090 GPUs. The fine-tuning for 20k iterations requires less than 6 hrs.

\PAR{Soft Attention-masking Scaling Factors.} In Sec.~\ref{subsec:encoder} of the main text, we explained our novel attention-masking mechanism which applies an additive offset to the $\text{Key}^T \text{Query}$ matrix. The offset is the mask value scaled by a positive scalar $\alpha$. We initialize $\alpha$ separately for each of the 8 attention heads as follows: [32, 32, 16, 16, 8, 8,  4, 4]. These are applied as learnable parameters which can be optimized by the network.

\PAR{Image Augmentations.} For the results reported in Table~\ref{tab:main_results} of the main text, we trained on image sequences generated by applying random affine transformations to COCO images. We use the popular $\smalltt{imgaug}$ library~\cite{imgaug} for this task. The range of values for each transformation type are as follows:

\begin{itemize} 
    \item Translation: $0-25\%$ w.r.t the dimension size.
    \item Rotation: $0-10\%$ in both directions.
    \item Shear: $0-10\%$ along both axes.
    \item Crop: $60 - 90\%$ of the image is retained.
\end{itemize}

Note that each image in the training image sequence is generated by applying the transformations to the original image, \ie we do not apply sequential augmentation. Aside from these geometric augmentations, we also apply color augmentations as follows:

\begin{itemize} 
    \item Hue : $0-12\%$
    \item Saturation: $0-12\%$
    \item Contrast (linear): $0-5\%$.
    \item Brightness: $0-25\%$
\end{itemize}

Our color augmentation strategy is inspired from that used by Cheng~\etal~\cite{Cheng21NeurIPS} for STCN.



\end{document}